# Tusom2021: A Phonetically Transcribed Speech Dataset from an Endangered Language for Universal Phone Recognition Experiments

*David R. Mortensen*[1]    *Jordan Picone*[ε]    *Xinjian Li*[1]    *Kathleen Siminyu*[χ]

[1]Carnegie Mellon University
[ε]University of Pittsburgh
[χ]Georgia Institute of Technology

`dmortens@cs.cmu.edu`

## Abstract

There is growing interest in ASR systems that can recognize phones in a language-independent fashion [1, 2, 3]. There is additionally interest in building langauge technologies for low-resource and endangered languages. However, there is a paucity of realistic data that can be used to test such systems and technologies. This paper presents a publicly available, phonetically transcribed corpus of 2255 utterances (words and short phrases) in the endangered Tangkhulic language East Tusom (no ISO 639-3 code), a Tibeto-Burman language variety spoken mostly in India. Because the dataset is transcribed in terms of phones, rather than phonemes, it is a better match for universal phone recognition systems than many larger (phonemically transcribed) datasets. This paper describes the dataset and the methodology used to produce it. It further presents basic benchmarks of state-of-the-art universal phone recognition systems on the dataset as baselines for future experiments.

## 1. Introduction

There have long been researchers who were interested in developing speech technologies for low-resource languages and in using speech technologies to document languages that are threatened or endangered. Recent advances have allowed the development of ASR systems that use multilingual (quasi-universal) phonetic models to perform phone recognition on arbitrary languages [2]. Such technology addresses the **transcription bottleneck** that hampers much documentary work on endangered languages. It has been estimated that it takes 40–100 hours of expert time to transcribe a single hour of audio in an endangered language [4]. Speech technologies that can relieve this bottleneck would be very valuable for language documentation, particularly if they could work without language-specific training data. Testing such technologies in a rigorous fashion, though, requires well-crafted datasets since the best such systems are not yet ready for field evaluation experiments. The goal of this paper is to introduce one such dataset, Tusom2021, a collection of trascribed audio from a comparative wordlist. The data is from the East Tusom language.

### 1.1. The East Tusom Language

East Tusom is a Tangkhulic (and therefore, Tibeto-Burman) language spoken in and around the easternmost of two villages in Manipur State, India, called "Tusom." It is one of many language varieties spoken by members of the Tangkhul ethnic group. These varieties, except for one, which serves as a lingua franca for the ethnic group, are conventionally called "dialects"

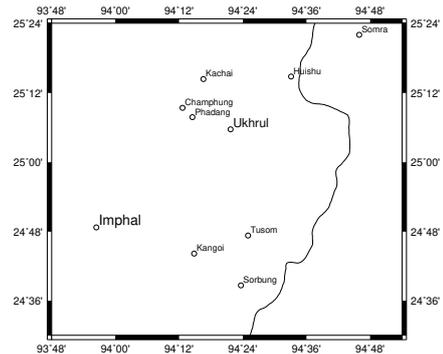

Figure 1: *Map of Southern Ukhrul District, Manipur State, India including East Tusom and neighboring towns (at the right of the map, the India-Myanmar border is shown)*

in English, though they display a remarkable level of variability. East Tusom differs from the lingua franca (Ukhrul or Standard Tangkhul) to a greater degree than German differs from Dutch. It is not known how many speakers of East Tusom there are, but reports indicate that there are fewer than 1,000. There is no writing system for this language and little documentation beyond first-draft transcriptions of (some of) the recordings from which the current resource was produced [5, 6].

The East Tusom language provides an interesting test case for universal ASR for a number of reasons:

- It is relatively rich phonetically, with a wide variety of consonant and vowel phones (Tables 1 and 2) as well as three tones (high, low, and falling).

- Some of these phones are not present in most of the world's languages, like front rounded vowels, back unrounded vowels, lateral fricatives, and voiceless trills [7].

- It is phonotactically interesting, with many relatively unusual consonant clusters like word-initial [sx] and [pχ] (see Table 3).

- East Tusom has 10 phonetic diphthongs (eight robustly attested, two marginal), some of which have proven challenging for our baseline phone recognition systems (see Table 4).

Furthermore, these particular data are useful because they realistically represent the quality and type of many of the recordings that are likely to be obtained in a field situation. The recording equipment was optimized for price and portability, rather than acoustic fidelity and there is considerable background noise (from air conditioning and small children).

|                              | Labial | Alveolar | Palatal | Velar | Uvular | Glottal |
|------------------------------|--------|----------|---------|-------|--------|---------|
| Plosive (voiced)             | b      | d        |         |       |        |         |
| Plosive (voiceless)          | p      | t        |         | k     |        | ʔ       |
| Plosive (voiceless aspirated)| pʰ     | tʰ       |         | kʰ    |        |         |
| Nasal Stop                   | m      | n        | ɲ       | ŋ     |        |         |
| Fricative (voiced)           | v      | z        | ʑ       |       |        |         |
| Fricative (voiceless)        | f      | s        | ɕ       | x     | χ      | h       |
| Lateral Fricative            |        | ɬ        |         |       |        |         |
| Trill (voiced)               |        | r        |         |       |        |         |
| Trill (voiceless)            |        | r̥        |         |       |        |         |
| Approximant (voiced)         |        | ɹ        |         |       |        |         |
| Approximant (voiceless)      |        | ɹ̥        | j       |       |        |         |
| Lateral Approximant          |        | l        |         |       |        |         |

Table 1: *Major consonant phones in East Tusom*

|      | front unrnded | rnded | central unrnded | rnded | back unrnded | rnded |
|------|---------------|-------|-----------------|-------|--------------|-------|
| high | i ĩ           | y ỹ   |                 |       | ɯ ɯ̃          | u ũ   |
| mid  | e ẽ           |       | ə ə̃             |       |              | o õ   |
| low  | (æ̃)           |       | a ã             |       |              |       |

Table 2: *Vowel phones in East Tusom*

|   | p  | t  | k  | b  | d | g |
|---|----|----|----|----|---|---|
| f | pf |    | kf |    |   |   |
| s | ps | ts |    |    |   |   |
| ɕ | pɕ | tɕ | kɕ |    |   |   |
| x | px | tx | kx |    |   |   |
| χ | pχ | tχ | kχ |    |   |   |
| v | pv |    | kv | bv |   |   |
| z | pz |    |    | bz |   |   |
| ʑ |    |    |    |    |   |   |

Table 3: *Initial consonant clusters in East Tusom*

### 1.2. The Task of Phone Recognition

It is possible to recognize different units of speech (that is, to recognize speech at different levels of representation). Words, BPE units, characters, and phonemes can all be recognized and each of them may be suited to particular use cases or applications. However, all of these are language-specific representations—a phoneme /p/ in English does not refer to the same linguistic and acoustic entity as the phoneme /p/ in Mandarin Chi-nese, for example. They are realized by different (allo)phones. Phone recognition seeks to optimize for universality by targeting units with relatively language-invariant acoustic and articulatory properties[1]. It refers to the task in which, given an acoustic signal, a corresponding sequence of phonetic characters—in the international phonetic alphabet (IPA) or some equivalent system like X-SAMPA—are generated.

| ∅ | i  | u  | e  | ə  | o    |
|---|----|----|----|----|------|
| i |    |    | ie | iə |      |
| y | yi |    |    |    |      |
| ɯ |    |    | ɯe | ɯə |      |
| u |    |    | ue | uə | (uo) |
| e |    |    |    |    |      |
| ə |    |    |    |    |      |
| o | oi |    |    |    |      |
| a |    | (au) |  |    |      |

Table 4: *East Tusom diphthongs*

### 1.3. Datasets for Phone Recognition

In past phone recognition experiments, a variety of different datasets have been used. [2] uses an uncorrected subset of the current Tusom dataset as well as a dataset comprised of transcriptions from various varieties of Inuktitut. The Inuktitut dataset is not publicly available for other experiments. The original Tusom dataset is too small, and of too low quality, to serve as a truly reliable test case for phone recognition. Recently, a dataset based on the UCLA Phonetics Archive has been produced [8]. This should prove useful as a test set for phone recognition as well as other low-resource ASR tasks. The UCLA dataset is about the same size as the current dataset and is more varied, but contains an unrealistically large number of uncommon phones since it was created from recordings meant to demonstrate the widest possible range of phonetic phenomena. It is also automatically aligned, resulting in some alignments which are not of gold quality. The Tusom2021 dataset is a useful complement to this dataset because it represents a realistic acoustic picture of the lexical vocabulary (mostly nouns and verbs) of a real, extremely low-resource, endangered language.

## 2. Methodology

Tusom2021 is based upon field recordings that were made in 2004 and several passes of transcription and correction starting in 2004 and continuing intermittently up to 2021, when a final version was completed. The data were collected from one native female speaker of East Tusom in her late 20s. She did not grow up in Tusom village but spoke the language at home, with her family, and when she stayed with her extended family in Tusom. Despite growing up outside of Tusom, she was extremely proficient in the language and very confident and consistent in her productions[2].

---

[1]True language in variance is probably not possible in an International Phonetic Alphabet (IPA)-oriented system. For example, voice onset time and the dimensions of the vowel space are continuous, but the IPA forces them into a discrete bed of Procrustes. This means that phones transcribed with these symbols will never be quite comparable.

The East Tusom speech was elicited based on a comparative word list[3] that was about 2,400 items in length. The word list, especially developed for languages in Northeast India, was designed to maximize the number of distinct roots, and therefore the phonetic diversity, of the elicited data.

Recording was carried out using an inexpensive lavalier microphone attached to the analog microphone jack of a laptop running the Audacity software package (under Microsoft Windows). Words were first elicited and transcribed with no recording. The first author, who had graduate training in phonetics and field linguistics, elicited an item from the speaker, then attempted to imitate it until the speaker confirmed that it was correctly produced, then transcribed it. The words were then divided into tranches. Each tranche was recorded in one session. For each item in each tranche, the first author provided an English word and the speaker responded with an East Tusom translation, repeated three times.

The transcriptions and the recordings were subsequently aligned manually, using Praat TextGrids [9]. The first author then made a pass over the transcriptions. Subsequently, the first author made a partial second pass, correcting transcriptions of tone. The second author made a global pass, checking all transcriptions for correctness and consistency. Final consistency checks were done using a combination of automatic and manual methods (using Python scripts to extract all tokens transcribed with a particular phone, then using auditory inspection and acoustic analysis—using Praat—to winnow out distinctions that had been inconsistently noted in the previous pass, such as voicing in plosives).

We estimate that the production of the transcriptions alone, leaving aside elicitation, recording, and segmentation, took over 200 hours.

## 3. Dataset

The Tusom2021 dataset[4] consists of 2393 tokens (each of which is a word or short phrase). For most types in the set, there are three tokens. In some cases, tokens of a type are not phonetically identical (and are not identically transcribed). The recordings are mono with a 22.05 kHz frame rate and they are distributed as WAV files. The cumulative length of the recordings is 51.8 minutes and they consists of just over 21,500 phones.

The transcriptions are mapped to the WAV files via a YAML file (`mapping.yml`). It consists of a large object the keys of which are file names and the values of which are objects with the following fields: `gloss` (the English gloss or translation of the token), `no_tones` (the phonetic transcription without any tone makers), `tone_dias` (the phonetic transcription with the tones indicated with combining diacritics), and `tone_letters` (the phonetic transcription with tones indicated using Chao tone letters after each syllable).

We anticipate that many users of this data will use the complete set to evaluate pretrained models. However, because some researchers may want to do experiments that rely on standardized train-dev-test splits, we have provided such splits. The standard splits are used in some of the experiments reported below. These partitions are defined in three YAML files with the same format as `mapping.yml`.

---

[2]But see the note on variation within utterance types below.
[3]English stimuli were mostly single words but East Tusom responses were often short phrases
[4]https://github.com/dmort27/tusom2021

| Set | Utterances | Phones | Minutes |
|---|---|---|---|
| train | 1578 | 15,132 | 36.5 |
| dev | 230 | 2,133 | 5.2 |
| test | 447 | 4,283 | 10.1 |
| all | 2255 | 21,548 | 51.8 |

Table 5: *Statistics for the Tusom2021 dataset (phone counts include tones)*

## 4. Experiments

### 4.1. Pretrained Models–Zero Shot

In this section, we evaluate two pretrained phone recognition models on the test set. Both models are open-sourced and available from GitHub[5]. The first model is a pretrained model (`eng`) which is trained using only English datasets: English Switchboard, English Tedlium and English LibriSpeech corpus [10, 11, 12]. The second model is a language-independent model (`lg_ind`) which is trained using many different languages. Both models use MFCCs as the input feature, bidirectional LSTMs as the encoder and CTC as the loss function [2, 13]. Each of the encoders has 5 layers and the hidden size of each layer is 640. For this experiment, we use the `no_tones` section of the Tusom dataset as both models cannot predict tones. We preprocess all Tusom utterances to build a phone inventory containing valid phones in this dataset. We also clean the dataset by normalizing rare phones whose frequency is less than 1% across the dataset. All those phones are replaced with a similar frequent phones in the inventory, the similarity measurement is done by PanPhon using phonological feature [14].

| set | model | PER | Add | Del | Sub |
|---|---|---|---|---|---|
| test | eng | 88.4% | 9.6% | 14.6% | 64.2% |
|  | lg_ind | 71.9% | 2.5% | 24.0% | 45.4% |
| all | eng | 88.9% | 9.8% | 14.7% | 64.3% |
|  | lg_ind | 69.2% | 3.2% | 23.7% | 42.4% |

Table 6: *Evaluation results on the test sets*

Table 6 shows the PER results for the two pretrained models as well as the categories of errors: addition (Add), deletion (Del), and substitution (Sub). The results for these two sets are strikingly similar, suggesting that the test set, while very small, is representative of the data set as a whole. In the test set, the English model has 88.4% PER and the language independent model improves the PER to 71.9%. The main reason for the gap between the English model and the language independent model is their inventory: the output of English model is English phonemes and it only covers a subset of the Tusom phones, on the other hand, the output of the language independent model is universal phone set and covers most of the Tusom phones. In the `all` set, we observe similiar trends and scores.

The most common substitution error in the English model is the ⟨kʰ, k⟩ pair, where the expected phone is [kʰ] but the model predicts [k] instead. This illustrates the limitation of model pretrained using a single language because [kʰ] does not appear among the phoneme labels for English and, even though it is a valid *phone* of English, a phoneme-based English model will not distinguish [k] from [kʰ]. However, in the language-independent model, both phones are covered in the model therefore both [kʰ] and [k] are available during inference. The most common substitution error in the language-independent model is the ⟨ə, a⟩ pair. Table 2 shows a typical example from the language-independent model. There are 2 substitution errors and 1 deletion in the utterance. While the error rate is as high as 60%, the decoded results is not very far from the reference phones.

---

[5]https://github.com/xinjli/allosaurus

Tusom2021 is clearly useful in distinguishing between phone recognition models in a zero-shot setting.

| Reference | kʰ | ə | m | ɯ | ə |
|---|---|---|---|---|---|
| | sub | sub | | | del |
| Hypothesis | k | a | m | ɯ | ε |

Figure 2: *An example of language independent results (East Tusom [kʰəmɯə] 'brother-in-law'; ε represents the empty string)*

### 4.2. Pretrained Models—Fine-Tuned

The first set of experiments compared zero-shot performance of two pretrained models, one trained on English and one trained on many languages in a largely language-independent fashion (Allosaurus) [2]. The second set of experiments, also using Allosaurus, explores how much using data from the Tusom2021 training set can improve performance through fine-tuning. We created training sets of different sizes: 10, 25, 50, 100, 250, 500, and 1000 utterances, as well as one using the entire training set of 1578 utterances. The sizes of the training sets were meant to correspond roughly to a geometric progression. We used the `no_tone` transcriptions from Tusom2021 (as in the first set of experiments). These transcriptions were segmented into phones according to the list of phones derived from the data. We fine tuned Allosaurus with each of the sets. In each instance, we set fine tuning to 250 epochs but the training took place over considerably fewer epochs due to early stopping. Results are given in Table 7. The best performance (PER of 33%) was obtained when tuning on the entire Tusom2021 test set for 20–30 epochs. This lowered PER from the baseline of 0.72 (see the first set of experiments) to 0.33. A plot showing PER for different number of epochs over the full range of data settings is given in 3. It shows that additional tuning examples have the greatest impact below 800 utterances. However, it appears that some (small) additional benefit would be derived from data beyond the 1,578 utterances in the Tusom2021 training set.

| | utterances | | | | | | | |
|---|---|---|---|---|---|---|---|---|
| epochs | 10 | 25 | 50 | 100 | 250 | 500 | 1000 | 1578 |
| 1 | 0.59 | 0.59 | 0.59 | 0.59 | 0.59 | 0.58 | 0.55 | 0.54 |
| 5 | 0.58 | 0.57 | 0.57 | 0.56 | 0.53 | 0.48 | 0.42 | 0.39 |
| 10 | 0.57 | 0.56 | 0.55 | 0.55 | 0.48 | 0.43 | 0.38 | 0.36 |
| 15 | 0.57 | 0.54 | 0.54 | 0.51 | 0.46 | 0.41 | 0.36 | 0.34 |
| 20 | 0.57 | 0.53 | 0.53 | 0.50 | 0.44 | 0.40 | 0.36 | **0.33** |
| 25 | 0.57 | 0.52 | 0.52 | 0.48 | 0.43 | 0.39 | 0.35 | **0.33** |
| 30 | 0.57 | 0.51 | 0.52 | 0.47 | 0.43 | 0.39 | 0.35 | **0.33** |
| 35 | 0.56 | 0.51 | 0.50 | 0.46 | 0.42 | 0.38 | 0.35 | 0.34 |
| 40 | 0.57 | 0.51 | 0.51 | 0.46 | 0.42 | 0.38 | 0.35 | 0.34 |
| 45 | 0.56 | 0.51 | 0.49 | 0.45 | 0.42 | 0.38 | 0.35 | 0.34 |

Table 7: *Results (phone error rate) from fine-tuning experiments*

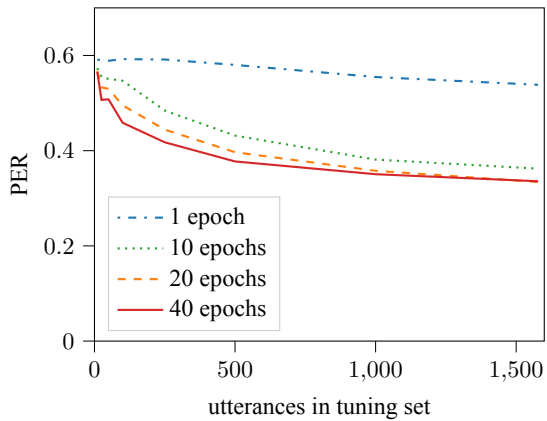

Figure 3: *PER as a function of tuning instances and epochs*

Table 8 shows the most common errors in the baseline that are corrected through fine-tuning. Unsurprisingly, the two most common corrected errors are vowel nasality and the presence of glottal stop, both of which can be acoustically ambiguous.

| reference | erroneous hypotheses | times corrected |
|---|---|---|
| ə | ə̃ | 17 |
| ʔ | ε | 8 |
| ʔ ɯ | ə | 7 |
| kʰ | k | 6 |
| i | e | 6 |
| õ | a | 6 |
| kʰ ə | k ə̃ | 5 |

Table 8: *Errors in baseline output corrected by fine-tuning on a larger quantity (1000 utterances) of data (ε is the empty string)*

These experiments illustrate two things about the Tusom2021 dataset: First, the transcriptions are consistent enough, and the training and test sets are representative enough, that tuning on the training set dramatically improves performance on the test set. Second, while the training set it small, it is large enough to fine-tune an Allosaurus model to near-optimal performance.

## 5. Conclusions

Tusom2021 addresses a significant lacuna in data resources for automatic phone recognition. However, it would be of even more value if it was one among many such datasets. A substantial group of documentary linguists are working to collect resources on the language of the word, particularly those that are threatened with extinction. Many of their projects produce phonemic transcriptions of corpora. It would be valuable, on several fronts, if some of these corpora could be transcribed phonetically (with sufficient phonetic realism that precise cross-linguistic comparisons are possible). As we have found, producing such a resource requires significant time and effort (at least 200 hours for less than 60 minutes of data, in our case) but results in a resource of enduring value.